\documentclass{article}
\usepackage{aaai}
\usepackage{times}
\usepackage{helvet}
\usepackage{courier}
\usepackage{url}
\usepackage{enumerate}

\usepackage{graphicx}
\graphicspath{{./Images/}}

\usepackage{moreverb}
\usepackage[normalem]{ulem}
\usepackage{amsmath,amsfonts}
\usepackage{amssymb}
\usepackage{bm}
\usepackage{mdwlist}

\usepackage[ruled,vlined,linesnumbered]{algorithm2e} 
\ifx \DontPrintSemicolon \undefined
 \def \DontPrintSemicolon {\dontprintsemicolon}
\fi
\newcommand{\acom}[1]{\tcc{\scriptsize #1}}
\usepackage{scalefnt}

\usepackage[usenames]{color} 

\newcommand{\commentout}[1]{}

\newcommand{\ie}{i.e.}
\newcommand{\eg}{e.g.}
\newcommand{\cf}{cf.}

\def\cS{\mathcal{S}}

\def\cO{\mathcal{O}}

\def\cA{\mathcal{A}}

\newcommand{\component}{\ensuremath{C}}
\newcommand{\rewardfn}{\ensuremath{r}}
\newcommand{\pivotingrewardfn}{\ensuremath{pr}}
\newcommand{\reward}{\ensuremath{R}}
\newcommand{\pivotingreward}{\ensuremath{pR}}
\newcommand{\pathreward}{\ensuremath{pathR}}
\newcommand{\start}{\ensuremath{*}}
\newcommand{\emptyfirewall}{\ensuremath{\emptyset}}

\ifx \theorem \undefined
\newtheorem{theorem}{Theorem}
\newtheorem{proposition}{Proposition}

\newtheorem{example}{Example}

{\hfill \rule[0.3ex]{1ex}{1ex} \par \addvspace{\bigskipamount}}

{\hfill \rule[0.3ex]{1ex}{1ex} \par \addvspace{\bigskipamount}}

\fi

\definecolor{Blue}{rgb}{0,0.16,0.90}
\definecolor{Red}{rgb}{0.90,0.16,0}
\definecolor{DarkBlue}{rgb}{0,0.08,0.45}
\definecolor{ChangedColor}{rgb}{0.9,0.08,0}

\newcommand{\pseudocode}[1]{}

\nocopyright

\title{POMDPs Make Better Hackers:\\ Accounting for Uncertainty in
  Penetration Testing}

\author{
  Carlos Sarraute\\
  Core Security \& ITBA\\
  Buenos Aires, Argentina\\
  \url{carlos@coresecurity.com}
  \And
  Olivier Buffet\\
  INRIA\\
  Nancy, France\\
  \url{buffet@loria.fr}
  \And
  J\"org Hoffmann\\
  Saarland University\\
  Saarbr\"ucken, Germany\\
  \url{hoffmann@cs.uni-saarland.de}
}

\begin{document}
\maketitle

\begin{abstract}
Penetration Testing is a methodology for assessing network security,
by generating and executing possible hacking attacks. Doing so
automatically allows for regular and systematic testing. A key
question is how to generate the attacks. This is naturally formulated
as planning under uncertainty, \ie, under incomplete knowledge about
the network configuration. Previous work uses classical planning, and
requires costly pre-processes reducing this uncertainty by extensive
application of scanning methods. By contrast, we herein model the
attack planning problem in terms of partially observable Markov
decision processes (POMDP). This allows to reason about the knowledge
available, and to intelligently employ scanning actions as part of the
attack. As one would expect, this accurate solution does not scale. We
devise a method that relies on POMDPs to find good attacks on
individual machines, which are then composed into an attack on the
network as a whole. This decomposition exploits network structure to
the extent possible, making targeted approximations (only) where
needed. Evaluating this method on a suitably adapted industrial test
suite, we demonstrate its effectiveness in both runtime and solution
quality.
\end{abstract}

\section{Introduction}
\label{introduction}

Penetration Testing (short {\em pentesting}) is a methodology for
assessing network security, by generating and executing possible
attacks exploiting known vulnerabilities of operating systems and
applications (e.g., \cite{ArcGra04}). Doing so automatically allows
for regular and systematic testing without a prohibitive amount of
human labor, and makes pentesting more accessible to non-experts. A
key question is how to automatically generate the attacks.

A natural way to address this issue is as an {\em attack planning}
problem. This is known in the AI Planning community as the ``Cyber
Security'' domain \cite{BodGohHaiHar05}. Independently (though
considerably later), the approach was put forward also by Core
Security \cite{LucSarRic10}, a company from the
pentesting industry. In that form, attack planning is very technical,
addressing the low-level system configuration details that are
relevant to vulnerabilities. %
Herein, we are concerned exclusively with this setting. We consider
regular automatic pentesting as done in Core Security's ``Core Insight
Enterprise'' tool. We will use the term ``attack planning'' in that
sense.

Lucangeli et al.\ \shortcite{LucSarRic10} encode attack planning into
PDDL, and use off-the-shelf planners. This already is useful---in
fact, it is currently employed commercially in Core Insight
Enterprise, using a variant of Metric-FF
\cite{hoffmann:jair-03}. However, the approach is limited by its
inability to handle uncertainty. The pentesting tool cannot be
up-to-date regarding all the details of the configuration of every
machine in the network, maintained by individual users.

Core Insight Enterprise currently addresses this by extensive use of
{\em scanning} methods as a pre-process to planning, which then
considers only {\em exploits}, \ie, hacking actions modifying the
system state. The drawbacks of this are that (a) this pre-process
incurs significant costs in terms of running time and network traffic,
and (b) even so, since scans are not perfect, a residual uncertainty
remains (Metric-FF is run based on the configuration that appears to
be most likely). Prior work \cite{SarRicLuc11} has addressed (b) by
associating each exploit with a success probability. This is unable to
model dependencies between the exploits, %
and it still requires extensive scanning (to obtain realistic success
probabilities) so does not solve (a). Herein, we provide the first
solution able to address both (a) and (b), intelligently mixing scans
with exploits like a real hacker would. The basic insight is that
penetration testing can be naturally modeled in terms of solving a
POMDP.

We encode the incomplete knowledge as an uncertainty of state, thus
modeling the possible network configurations in terms of a probability
distribution. Scans and exploits are deterministic in that their
outcome depends only on the state they are executed in. Negative
rewards encode the cost (the duration) of scans and exploits; positive
rewards encode the value of targets attained. The model incorporates
firewalls, detrimental side-effects of exploits (crashing programs or
entire machines), %
and dependencies between exploits relying on similar vulnerabilities.

POMDP solvers fail to scale to large networks. This is not
surprising---even the input model grows exponentially in the number of
machines. We show how to address this based on exploiting network
structure. We view networks as graphs whose vertices are
fully-connected subnetworks, and whose arcs encode the connections
between these, filtered by firewalls.  We decompose this graph into
biconnected components. We approximate the attacks on these components
by combining attacks on individual subnetworks. We approximate the
latter by combining attacks on individual machines. The approximations
are conservative, \ie, they never over-estimate the value of the
policy returned. %
Attacks on individual machines are modeled and solved as POMDPs, and
the solutions are propagated back up. We evaluate this approach based
on the test suite of Core Insight Enterprise, showing that, compared
to a global POMDP model, it vastly improves runtime at a small cost in
attack quality.

We next discuss some preliminaries. We then describe our POMDP model,
our decomposition algorithm, and our experimental findings, before
concluding the paper.

\section{Preliminaries}
\label{preliminaries}

We fill in some details on network structure and penetration
testing. We give a brief background on POMDPs.

\subsection{Network Structure}

Networks can be viewed as directed graphs whose vertices are given by
the set $M$ of \emph{machines}, and whose arcs are connections between
pairs of $m \in M$. However, in practice, these network graphs have a
particular structure. They tend to consist of \emph{subnetworks}, \ie,
clusters $N$ of machines where every $m \in N$ is directly connected
to every $m' \in N$. %
By contrast, not every subnetwork $N$ is connected to every other
subnetwork $N'$, and typically, if such a connection does exist, then
it is filtered by a \emph{firewall}.

From the perspective of an attacker, the firewalls filter the
connections and thus limit the attacks that can be executed when
trying to hack into a subnetwork $N'$ from another subnetwork $N$.  On
the other hand, once the hacker managed to get into a subnetwork $N$,
access to all machines within $N$ is easy. Thus a natural
representation of the network, from an attack planning point of view,
is that of a graph whose vertices are subnetworks, and whose arcs are
annotated with firewalls $F$. We herein refer to this graph as the
\emph{logical network} $LN$, and we denote its arcs with $N
\xrightarrow{F} N'$.

We formalize firewalls as sets of rules describing which kinds of
communication (\eg, ports) are disallowed. Thus smaller sets
correspond to ``weaker'' firewalls, and the \emph{empty firewall}
blocks no communication at all.

We remark that, in our POMDP model, we do not provide for privilege
escalation, or obtaining passwords. This can instead be modeled at the
level of $LN$. Different privilege levels on the same machine $m$ can
be encoded via different copies of $m$. If controlling $m$ allows
the retrieval of passwords, then $m$ can be connected via empty firewalls
to the machines $m'$ who can be accessed by using these passwords, more
precisely to high-privilege copies of these $m'$.

\subsection{Penetration Testing}

Uncertainty in pentesting arises because it is impossible to keep
track of all the \emph{configuration} details of individual machines,
i.e., exactly which versions of which programs are installed
etc. However, it is safe to assume that the pentesting tool knows the
structure of the network, \ie, the graph $LN$ and the filtering done
by each firewall: changes to this are infrequent and can easily be
registered.

The objective of pentesting is to gain control over certain machines
(with critical content) in the network. At any point in time, each
machine has a unique {\em status}. A \emph{controlled} machine $m$ has
already been hacked into. A \emph{reached} machine $m$ is connected to
a controlled machine, \ie, either $m$ is in a subnetwork $N$ one of
whose machines is controlled, or $m$ is in a subnetwork $N'$ with a
$LN$ arc $N \xrightarrow{F} N'$ where one of the machines in $N$ is
controlled. All other machines are \emph{not reached}. The algorithm
starts with one controlled machine, denoted here by
\start.\footnote{For simplicity, we will notate \start\ as a separate
  vertex in $LN$.  If \start\ is part of a subnetwork $N$, this means
  to turn $N \setminus \{\start\}$ into a separate vertex in $LN$,
  connected to \start\ via the empty firewall.}
We will use the following (small but real-life) situation as a running
example:

\vspace{-0.05cm}
\begin{example}\label{exp:running-preliminaries}
The attacker has already hacked into a machine $m'$, and now wishes to
attack a machine $m$ within the same subnetwork. The attacker knows
two exploits: \emph{SA}, the ``Symantec Rtvscan buffer overflow
exploit''; and \emph{CAU}, the ``CA Unicenter message queuing
exploit''. SA targets a particular version of ``Symantec Antivirus'',
that usually listens on port 2967. CAU targets a particular version of
``CA Unicenter'', that usually listens on port 6668. Both work only if
a protection mechanism called \emph{DEP} (``Data Execution
Prevention'') is disabled.
\end{example}
\vspace{-0.05cm}

If SA fails, then it is likely that CAU will fail as well (because DEP
is enabled). The attacker is then better off trying something
else. Achieving such behavior requires the attack plan to observe the
outcomes of actions, and to react accordingly. Classical planning
(which assumes perfect world knowledge at planning time) cannot
accomplish this.  

Furthermore, port scans---observation actions testing whether or not a
particular port is open---should be used only if one actually intends
to execute a relevant exploit. Here, if we start with SA, we should
scan only port 2967. We accomplish such behavior through the use of
POMDPs. By contrast, to reduce uncertainty, classical planning
requires a pre-process executing \emph{all} possible scans. In this
example, there are only two---ports 2967 and 6668---however in general
there are many, causing significant network traffic and waiting time.

\subsection{POMDPs}

POMDPs are usually defined (e.g., \cite{Monahan82,KaeLitCas-aij98}) by
a tuple $\langle \cS, \cA, \cO, T, O, r, b_0 \rangle$. If the system
is in state $s \in \cS$ (the {\em state space}), and the agent
performs an action $a\in \cA$ (the {\em action space}), then that
results in (1) a transition to a state $s'$ according to the {\em
  transition function} $T(s,a,s')=Pr(s'| s,a)$, (2) an observation
$o\in\cO$ (the {\em observation space}) according to the {\em
  observation function} $O(s',a,o)=Pr(o| s',a)$ and (3) a scalar {\em
  reward} $r(s,a,s')$. $b_0$, the {\em initial belief}, is a
probability distribution over $\cS$. %

The agent must find a decision {\em policy} $\pi$ choosing, at each
step, the best action based on its past observations and actions so as
to maximize its future gain, which we measure here through the total
accumulated reward. The expected value of an optimal policy is denoted
with $V^*$.

The agent typically reasons about the hidden state of the system using
a {\em belief state} $b$, a probability distribution over $\cS$. For
our experiments we use SARSOP \cite{KurHsuLee08}, a state of the art
point-based algorithm, i.e., an algorithm approximating the value
function as the upper envelope of a set of hyperplanes, corresponding
to a selection of particular belief states (referred to as
``points'').

\section{POMDP Model}
\label{POMDP-model}

A preliminary version of our POMDP model appeared at the SecArt'11
workshop \cite{SarBufHof11}. The reader may refer to that paper for a
more detailed example listing complete transition and observation
models for some actions, and exemplifying the evolution of belief
states when applying these actions. In what follows, we keep the
description brief in the interest of space.

\subsection{States}
\label{POMDP-model:states}

Several aspects of the problem---notably the network structure and the
firewall filtering rules---are known and static. POMDP variables
encoding these aspects can be compiled out in a pre-process, and are
not included in our model.

The states capture the status of each machine (controlled/reached/not
reached). For non-controlled machines, they also specify the software
configuration (operating system, servers, open ports, \dots). We
specify the vulnerable programs, as well as programs that can provide
information about these (e.g., the protection mechanism ``DEP'' in our
running example is relevant to both exploits). The states also
indicate whether a given machine or program has crashed. 

Finally, we introduce one special \emph{terminal} state into the POMDP
model (of the entire network, not of individual machines). That state
corresponds to giving up the attack, when for every available action
(if any) the potential benefit is not worth the action's cost.

\vspace{-0.05cm}
\begin{example}\label{exp:running-states}
The states describe the attacked machine $m$. For simplicity, we
assume that the exploits here do not risk crashing the machine (see
also next sub-section). Apart from the terminal state and the state
representing that $m$ is controlled, the states specify which programs
(``SA'' or ``CAU'') are present, whether they are vulnerable, and
whether ``DEP'' is enabled. Each application is listening on a
different port, so a port is open iff the respective application is
present, and we do not need to model ports separately. Thus we have a
total of $20$ states:\\ {\scriptsize
\begin{minipage}[t]{.24\linewidth}
\begin{verbatim} 
1 terminal
2 m_controlled
\end{verbatim}
\end{minipage}
\begin{minipage}[t]{.34\linewidth}
\begin{verbatim}
 3 m_none
 4 m_CAU
 5 m_CAU_Vul
 6 m_SA
 7 m_SA_CAU
 8 m_SA_CAU_Vul
 9 m_SA_Vul
10 m_SA_Vul_CAU
11 m_SA_Vul_CAU_Vul
\end{verbatim}
\end{minipage}
\begin{minipage}[t]{.30\linewidth}
\begin{verbatim}
12 m_DEP_none
13 m_DEP_CAU
14 m_DEP_CAU_Vul
15 m_DEP_SA
16 m_DEP_SA_CAU
17 m_DEP_SA_CAU_Vul
18 m_DEP_SA_Vul
19 m_DEP_SA_Vul_CAU
20 m_DEP_SA_Vul_CAU_Vul
\end{verbatim}
\end{minipage}
}
\end{example}

In short, the states for each machine $m$ essentially are tuples of
status values for each relevant program. Global system states then are
tuples of these machines-states, with one entry for each $m \in
M$. The state space enumerates these tuples. In other words, the state
space is factored in a natural way, by programs and machines. An
obvious option is, thus, to model and solve the problem using factored
POMDPs (e.g., \cite{HanFen00}). We did not try this yet; our POMDP
model generator internally enumerates the states, and feeds the ground
model to SARSOP.\footnote{Note that this approach enables certain
  non-trivial optimizations: some of the states in
  Example~\ref{exp:running-states} could be merged. If DEP is enabled,
  then it does not matter whether or not CAU/SA are vulnerable. For
  brevity, we do not discuss this in detail here.}

The factored nature of our problem also implies that the state space
is huge. In a realistic setting, the set $C$ of possible configuration
tuples for each machine $m \in M$ is very large, yielding an enormous
state space $|\cS|=O(|C|^{|M|})$. In practice, we will run POMDPs only
on single machines, i.e., $|M|=1$.

\subsection{Actions}
\label{POMDP-model:actions}

To reach the terminal state, we need a \emph{terminate} action
indicating that one gives up on the attack. 

There are two main types of actions, \emph{scans} and \emph{exploits},
which both have to be targeted at reachable machines.
Scans can be OS detection actions or port scans. In most cases, they
have no effect on the state of the target machine. Their purpose is to
gain knowledge about a machine's configuration, by an observation that
typically allows to prune some states from the belief (e.g., observing
that the OS must be some Windows XP version). %
Exploits make use of a vulnerability---if present---to gain control
over a machine. The outcome of the exploit is observed by the
attacker, so a failed exploit may, like a scan, yield information
about the configuration (e.g., that a protection mechanism is likely
to be running). For a minority of exploits, a failed attempt crashes
the machine. 

For all actions, the outcome is deterministic: which observation is
returned, and whether an exploit succeeds/fails/crashes, is uniquely
determined by the target machine's configuration.

\begin{example}\label{exp:running-actions}
In our example, there are five possible actions:\\ {\scriptsize
\begin{minipage}[t]{.4\linewidth}
\begin{verbatim}
m_exploit_SA
m_exploit_CAU
m_scan_port_2967
m_scan_port_6668
terminate
\end{verbatim}
\end{minipage}
}

\smallskip
\noindent
The POMDP model specifies, for each state in
Example~\ref{exp:running-states}, the outcome of each action. For
example, {\scriptsize \verb+m_exploit_SA+} succeeds if and only if SA
is present and vulnerable, and DEP is disabled. Hence, when applied to
either of the states 9, 10, or 11, {\scriptsize \verb+m_exploit_SA+}
results in state 2, and returns the observation {\scriptsize
  \verb+succeeded+}. Applied to any other state, {\scriptsize
  \verb+m_exploit_SA+} leaves the state unchanged, and the observation
is {\scriptsize \verb+failed+}.
\end{example}

The outcomes of actions also depend on what firewall (if any) stands
between the pentester and the target. If the firewall filters out the
relevant port, then the action is unusable: its transition model
leaves the state unchanged, and no observation is returned. For
example, if a firewall $F$ filters out port 2967, then {\scriptsize
  \verb+m_scan_port_2967+} and {\scriptsize \verb+m_exploit_SA+} are
unusable through $F$, but can be employed as soon as a machine behind
$F$ is under control.

\subsection{Rewards}
\label{POMDP-model:rewards}

No reward is obtained when using the \emph{terminate} action or when
in the terminal state. 

The instant reward of any scan/exploit action depends on the
transition it induces in the present state. Our simple model is to
additively decompose the instant reward $\rewardfn(s,a,s')$ into
$\rewardfn(s,a,s') = r_e(s,a,s') + r_t(a) + r_d(a)$.  Here, (i)
$r_e(s,a,s')$ is the value of the attacked machine in case the
transition $(s,a,s')$ corresponds to a successful exploit, and is $0$
for all other transitions; (ii) $r_t(a)$ is a cost that depends on the
action's duration; %
and (iii) $r_d(a)$ is a cost that reflects the risk of detection when
using this action. (iii) is orthogonal to the risk of crashing a
program/machine, which as described we model as a possible outcome of
exploits. Note that (ii) and (iii) may be correlated; however, there
is no 1-to-1 correspondence between the duration and detection risk of
an exploit, so it makes sense to be able to distinguish these
two. Finally, note that (i) results in summing up rewards for
successful exploits on different machines. That is not a limiting
assumption: one can reward breaking into \emph{[$m_1$ OR $m_2$]} by
introducing a new virtual machine, accessible at no cost from each of
$m_1$ and $m_2$.

\vspace{-0.05cm}
\begin{example}\label{exp:running-rewards}
In our example, we set $r_e=100$ in case of success, $0$ otherwise;
$r_t=-10$ for all actions; and $r_d=0$ (no risk of detection). We will
see below what effect these settings have on an optimal policy.
\end{example}
\vspace{-0.05cm}

Since all actions are deterministic, there is no point in repeating
them on the same target through the same firewall---this will not
produce new effects or bring any new information. In particular,
positive rewards cannot be received multiple times. Thus cyclic
behaviors incur infinite negative costs. This implies that the
expected reward of an optimal policy is finite even without
discounting.\footnote{In fact, the problem falls into the class of
  \emph{Stochastic Shortest Path Problems} \cite{BerTsi96}.}

\subsection{Designing the Initial Belief}
\label{POMDP-model:initial-belief}

Penetration testing is done at regular time intervals. The initial
belief---our knowledge of the network when we start the
pentesting---depends on (a) what was known at the end of the previous
pentest, and on (b) what may have changed since then. We assume for
simplicity that knowledge (a) is perfect, i.e., each machine $m$ at
time $0$ (the last pentest) is assigned one concrete configuration
$I(m)$. We then compute the initial belief as a function $b_0(I,T)$
where $T$ is the number of days elapsed since the last pentest. The
uncertainty in this belief arises from not knowing which software
updates were applied. We assume that the updates are made
independently on each machine (simplifying, but reasonable given that
updates are controlled by individual users).

A simple model of updates \cite{SarBufHof11} encodes the uncertain
evolution of each program independently, in terms of a Markov
chain. The states in each chain correspond to the different versions
of the program, and the transitions model the possible program updates
(with estimated probabilities that these updates will be made). The
initial belief then is the distribution resulting from this chain
after $T$ steps.

\vspace{-0.05cm}
\begin{example}\label{exp:generating-initial-belief}

  \begin{figure}[t]

    \begin{minipage}{0.3\linewidth}
      \centerline{\includegraphics[width=0.95\linewidth]{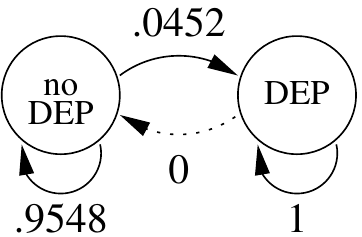}}
    \end{minipage}
    \hfill
    \begin{minipage}{0.3\linewidth}
      \centerline{\includegraphics[width=0.95\linewidth]{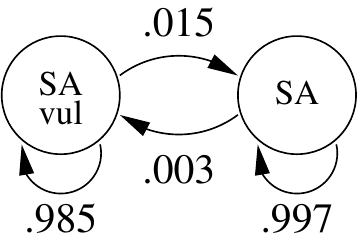}}
    \end{minipage}
%
\hfill
      \begin{minipage}{0.3\linewidth}
        \centerline{\includegraphics[width=0.95\linewidth]{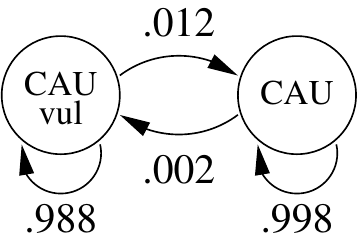}}
      \end{minipage}
\vspace{-0.2cm}
    \caption{The three independent Markov chains used to model the
      update mechanism in our example network.}
    \label{fig:MarkovChains}
\vspace{-0.6cm}
  \end{figure}

  In our running example, the three components in the single machine
  are DEP, CAU and SA. They are updated via three independent Markov
  chains, each with two states, as illustrated in
  Figure~\ref{fig:MarkovChains}. The probabilities indicate how likely
  the machine is to transition from one state to another during one
  day. Say we set $T=30$, and run the Markov chains on the
  configuration $I$ in which $m$ has DEP disabled, and both SA and CAU
  are vulnerable to the attacker's exploit. In the resulting initial
  belief $b_0(I,T)$, DEP is likely to be enabled; the weight of states
  12--20 in Example~\ref{exp:running-states} is high in $b_0$
  ($>70\%$).
\end{example}
\vspace{-0.05cm}

Here, we use this simple model as the basic building block in a method
taking into account that version $x$ of program A may need version $y$
or $z$ of program B. We assume that programs are organized in a
hierarchical manner, the operating system being at the root of a
directed acyclic graph, and a program having as its parents the
programs it directly depends on. This yields a Dynamic Bayesian
Network, where each conditional probability distribution is derived
from a Markov chain $Pr(X_t=x'|X_{t-1}=x)$ %
filtered by a compatibility function $\delta(X=x,parent_1(X)=x_1,
\dots, parent_k(X)=x_k)$, that returns $1$ iff the value of $X$ is
compatible with the parent versions, $0$ otherwise. This model of
updates is reasonable, but of course still not realistic; future work
needs to investigate such models in detail.

We now illustrate how reasoning with the probabilities of the initial
belief results in the desired intelligent behavior.

\vspace{-0.05cm}
\begin{example}\label{exp:running-initial-belief-policy}
Say we compute the initial belief $b_0(I,T)$ as in
Example~\ref{exp:generating-initial-belief}. Since the weight of
states 12--20 is high in $b_0$, if {\scriptsize \verb+m_exploit_SA+}
fails, then the success probability of {\scriptsize
  \verb+m_exploit_CAU+} is reduced to the point of not being worth the
effort anymore, and the attacker (the optimal policy) gives up, i.e.,
would try a different attack not prevented by DEP. Namely, consider
$Pr($CAU$^+|$2967$^+)$, i.e., the probability of {\scriptsize
  \verb+m_exploit_CAU+} succeeding, after observing that port 2967 is
open. This corresponds to the weight of (A) states 8 and 11 in
Example~\ref{exp:running-states}, within the states (B) 6--11 plus
15--20. That weight (A/B) is about $20\%$. Thus the expected value of
{\scriptsize \verb+m_exploit_CAU+} in this situation is about
$100*0.2$ [success reward] $- 10$ [action cost] $= 10$,
cf.\ Example~\ref{exp:running-rewards}, so the action is
worthwhile. By contrast, say that {\scriptsize \verb+m_exploit_SA+}
has been tried and failed. Then (A) is reduced to state 8 only, while
(B) still contains (in particular) all the DEP states 15--20. The
latter states have a lot of weight, and thus
$Pr($CAU$^+|$2967$^+,$SA$^-)$ is only about $5\%$. Given this, the
expected value of {\scriptsize \verb+m_exploit_CAU+} is negative, and
it is better to apply {\scriptsize \verb+terminate+} instead.
\end{example}
\vspace{-0.05cm}

\section{4AL Decomposition Algorithm}
\label{decomposition-algorithm}

\ifthenelse{\isundefined{\jfpda}}{
\begin{figure*}[t]
\begin{tabular}{ccc}
    \scalebox{0.7}{
      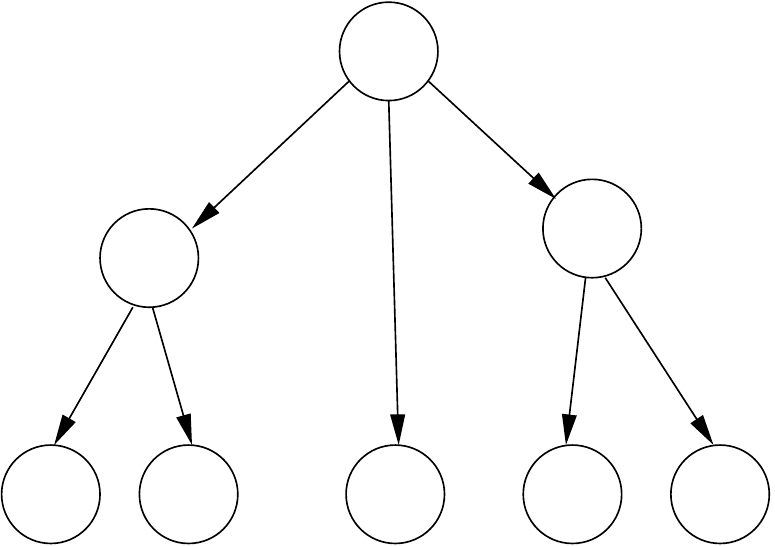
    } &
    \scalebox{0.7}{
      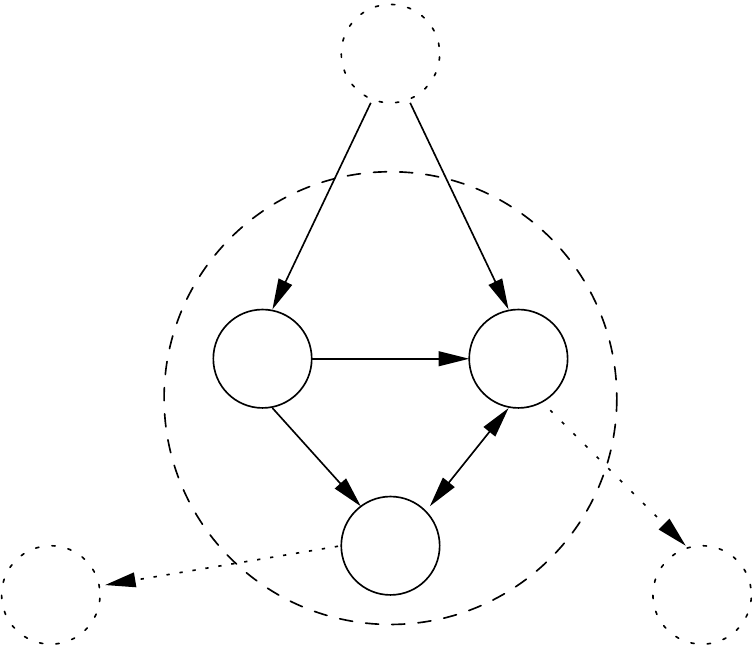
    } &
    \scalebox{0.7}{
      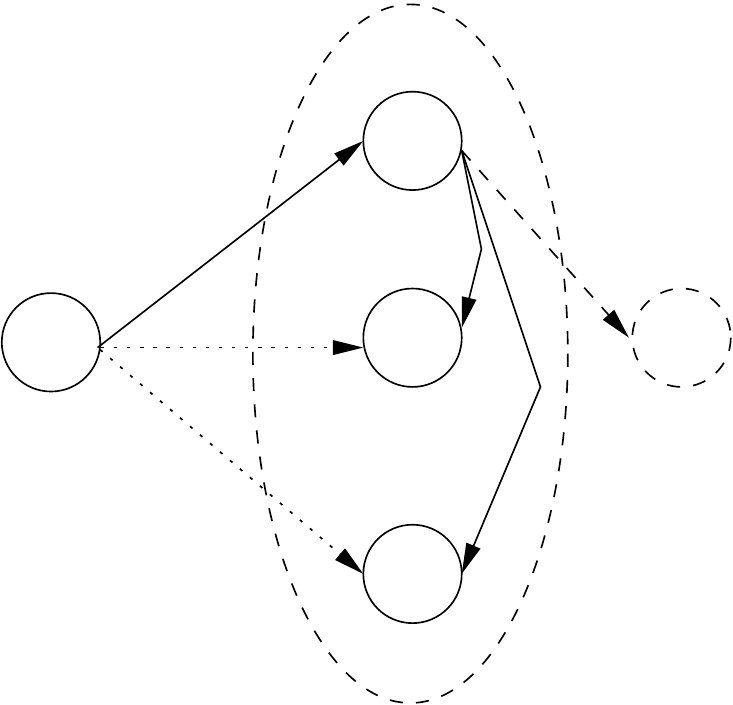
    } \\[0.1cm]
    (a) $LN$ as tree of components $C$. & 
    (b) Paths for attacking $C_1$. & 
    (c) Attacking $N_3$ from $N_1$, using $m$ first.   
  \end{tabular}
  \vspace{-0.35cm}
  \caption{Illustration of Levels 1, 2, and 3 (from left to right) of
    the 4AL algorithm.}
  \label{fig:4ALlevels123}
  \vspace{-0.5cm}
\end{figure*}
}{
  \begin{figure}[t]
    \begin{minipage}{0.45\linewidth}
      \centering
      \scalebox{0.7}{
        \input{Images/4AL1.pdf_t}
      }
      \\
      (a) $LN$ as tree of components $C$.
    \end{minipage}
    \hfill  
    \begin{minipage}{0.45\linewidth}
      \centering
      \scalebox{0.7}{
        \input{Images/4AL2.pdf_t}
      }
      \\
      (b) Paths for attacking $C_1$.
    \end{minipage}

    \medskip
    
    \begin{minipage}{1\linewidth}
      \centering
      \scalebox{0.7}{
        \input{Images/4AL3.pdf_t}
      }
      \\
      (c) Attacking $N_3$ from $N_1$, using $m$ first.   
    \end{minipage}
  \caption{Illustration of Levels 1, 2, and 3 (from left to right) of
    the 4AL algorithm.}
  \label{fig:4ALlevels123}
\end{figure}
}

\begin{figure*}[t]
\centering
\begin{tabular}{cc}
\input{Algorithms/4AL1}
&
\input{Algorithms/4AL2}
\\
\input{Algorithms/4AL3}
&
\input{Algorithms/4AL4}
\end{tabular}
\vspace{-0.2cm}\caption{4AL algorithm, pseudo-code.}
\label{fig:4ALcode}
\vspace{-0.5cm}
\end{figure*}

As hinted, POMDPs do not scale to large networks (cf.\ the experiments
in the next section). We now present an approach using decomposition
and approximation to overcome this problem, relying on POMDPs only to
attack individual machines. The approach is called \emph{4AL} since it
addresses network attack at 4 different levels of abstraction. 4AL is
a POMDP solver specialized to attack planning as addressed here. Its
input are the logical network $LN$ and POMDP models encoding attacks
on individual machines. Its output is a policy (an attack) for the
global POMDP encoding $LN$, as well as an approximation of the value
of the global value function. We next overview the algorithm, then
fill in some technical details. To simplify the presentation, we will
focus on the approximation of the value function, and outline only
briefly how to construct the policy.

\subsection{4AL Overview and Basic Properties}
\label{decomposition-algorithm:overview}

The four levels of 4AL are: (1) \emph{Decomposing the Network}, (2)
\emph{Attacking Components}, (3) \emph{Attacking Subnetworks}, and (4)
\emph{Attacking Individual Machines}. We outline these levels in
turn before providing technical details. Figure~\ref{fig:4ALlevels123} provides illustrations.

\vspace{-0.05cm}
\begin{itemize}
\item \textbf{Level~1:} Decompose the logical network $LN$ into a tree
  of biconnected components, rooted at \start. In reverse topological
  order, call Level~2 on each component; propagate the outcomes
  upwards in the tree.
\end{itemize}
\vspace{-0.05cm}

Every graph decomposes into a unique tree of biconnected components
\cite{HopTar-cACM73}. A biconnected component is a sub-graph that
remains connected when removing any one vertex. In pentesting,
intuitively this means that there is more than one possibility (more
than one path) to attack the subnetworks within the component,
requiring to reason about the component as a whole (which is the job
of Level~2). By contrast, if removing subnetwork $X$ (\eg, $N_2$
in Figure~\ref{fig:4ALlevels123} (b)) makes the graph fall apart into
two separate sub-graphs ($C_2$ vs.\ the rest of $LN$, compare also
Figure~\ref{fig:4ALlevels123} (a)), then \emph{all} attacks from
\start\ to one of these sub-graphs ($C_2$ here) must first traverse
$X$ ($N_2$ here). Thus the overall expected value of the attack can be
computed by (1) computing the value of attacking that sub-graph
($C_2$) alone, %
and (2) adding the result as a \emph{pivoting reward} to the reward of
breaking into $X$ ($N_2$). In other words, we ``propagate the outcomes
upwards'' in the tree displayed in Figure~\ref{fig:4ALlevels123}
(a). 

It is important to note that this tree decomposition will typically
result in a huge reduction of complexity. Biconnected components in
$LN$ arise only from clusters of more than $2$ subnetworks sharing a
common (physical) firewall machine. Such clusters tend to be small. In
the real-world test scenario used by Core Security and in our
experiment here, there is only one cluster, of size $3$. In case there
are no clusters at all, $LN$ is a tree and 4AL Level~2 trivializes
completely.

\vspace{-0.05cm}
\begin{itemize}
\item \textbf{Level~2:} Given component $\component$, consider, for
  each rewarded subnetwork $N \in \component$, all paths $P$ in
  $\component$ that reach $N$. Backwards along each $P$, call Level~3
  on each subnetwork and associated firewall. Choose the best path for
  each $N$. Aggregate these path values over all $N$, by summing up
  but disregarding rewards that were already accounted for by a
  previous path in the sum.
\end{itemize}
\vspace{-0.05cm}

In case a biconnected component $\component$ contains more than one
subnetwork, to obtain the best attack on $\component$, in general we
have no choice but to encode the entire component as a POMDP. Since
that is not feasible, Level~2 considers individual ``attack paths''
within $\component$. Any single path $P$ is equivalent to a sequence
of attacks on individual subnetworks; these attacks are evaluated
using Level~3. We consider the rewarded vertices $N$ in separation,
enumerating the attack paths and choosing a best one.  The values of
the best paths are aggregated over all $N$ in a conservative
(pessimistic) manner, by accounting for each reward at most once. A
strict under-estimation occurs in case the best paths for some
rewarded vertices are not disjoint: then these attacks share some of
their cost, so a combined attack has a higher expected reward than the
sum of independent attacks.

In Figure~\ref{fig:4ALlevels123} (b), $N_2$ and $N_3$ have a pivoting
reward because they allow to reach the components $C_2$ and $C_3$
respectively. If the best paths for both $N_2$ and $N_3$ go via $N_1$
(because the firewall $F^*_3$ is very strict), then these paths are
not disjoint, duplicating the effort for breaking into $N_1$.

Obviously, enumerating attack paths within $\component$ is exponential
in the size of $\component$. This is the only point in 4AL---apart of
course from calls to the POMDP solver---that has worst-case
exponential runtime. In practice, biconnected components are typically
small, \cf\ the above.

\vspace{-0.05cm}
\begin{itemize}
\item \textbf{Level~3:} Given a subnetwork $N$ and a firewall $F$
  through which to attack $N$, for each machine $m \in N$ approximate
  the reward obtained when attacking $m$ first. For this, modify $m$'s
  reward to take into account that, after breaking $m$, we are behind
  $F$: call Level~4 to obtain the values of all $m' \neq m$ with an
  empty firewall; then add these values, plus any pivoting reward, to
  the reward of $m$ and call Level~4 on this modified $m$ with
  firewall $F$. Maximize the resulting value over all $m \in N$.
\end{itemize}
\vspace{-0.05cm}

Consider Figure~\ref{fig:4ALlevels123} (c). When attacking $N$ (here,
$N_3$) from some machine behind the firewall $F$ (here, $F^1_3$), we
have to choose which machine inside $N$ to attack. Given we commit to
one such choice $m$, the attack problem becomes that of breaking into
$m$ and afterwards exploiting the direct connection to any $m \neq m'
\in N$, and any descendant network (here, $C_3$) we can now pivot
to. As described, that can be dealt with by combining attacks on
individual machines with modified rewards. (The pivoting reward for
descendant networks is computed beforehand by Levels 1 and 2.)

Like Level~2, Level~3 makes a conservative approximation. It fixes a
choice of which $m \in N$ to attack. By contrast, the best strategy
may be to switch between different $m \in N$ depending on the success
of the attack so far. For example, if one exploit is very likely to
succeed, then it may pay off to try this on all $m$ first, before
trying anything else.

\vspace{-0.05cm}
\begin{itemize}
\item \textbf{Level~4:} Given a machine $m$ and a firewall $F$, model
  the single-machine attack planning problem as a POMDP, and run an
  off-the-shelf POMDP solver. Cache known results to avoid duplicate
  effort.
\end{itemize}
\vspace{-0.05cm}

This last step should be self-explanatory. The POMDP model is created
as described earlier. Note that Level~3 may, during the execution of
4AL, call the same machine with the same firewall more than once. For
example, in Figure~\ref{fig:4ALlevels123} (c), when we switch to
attacking $m'_1$ instead of $m$, the call of Level~4 with $m'_k$ and
an empty firewall is repeated.

Summing up, 4AL has low-order polynomial runtime except for the
enumeration of paths within biconnected components (Level~2), and
solving single-machine POMDPs (Level~4). The decomposition at Level~1
incurs no information loss. Levels 2 and 3 make conservative
approximations, so, if the POMDP solutions are conservative (\eg,
optimal), then the overall outcome of 4AL is conservative as well.

\subsection{Technicalities}
\label{decomposition-algorithm:technicalities}

To provide a more detailed understanding of 4AL, we now discuss
pseudo-code for the algorithm, provided in
Figure~\ref{fig:4ALcode}. Consider first
Algorithm~\ref{alg:level1}. It should be clear how the overall
structure of the algorithm corresponds to our previous discussion. It
calls the linear-time algorithm by Hopcroft and Tarjan
\shortcite{HopTar-cACM73} (hereafter, HT) to find the
decomposition. The loop $i=k,\dots,1$ processes the components in
reverse topological order. The pivoting reward function
$\pivotingrewardfn$ encodes the propagation of rewards upwards in the
tree; this should be self-explanatory apart for the expression ``the
parent'' of $C_i$ in $LN$. The latter relies on the fact that, after
``clean-up'' (line 2), each component has exactly one such parent.

To explain the clean-up, note first that HT works on undirected
graphs; when applying it, we ignore the direction of the arcs in
$LN$. The outcome is an undirected tree of biconnected components,
where the \emph{cut vertices}---those vertices removing which makes
the graph break apart---are shared between several components. In
Figure~\ref{fig:4ALlevels123} (b), \eg, $N_2$ prior to the clean-up
belongs to both, $C_1$ and $C_2$. The clean-up sets the root of the
tree to \start, and assigns each cut-vertex to the component closest
to \start\ (\eg, $N_2$ is assigned to $C_1$); \start\ itself is turned
into a separate component. Re-introducing the direction of arcs in
$LN$, we then prune vertices not reachable from \start. Next, we
remove arcs that cannot participate in any non-redundant attack path
starting in \start. Since moving \emph{towards} \start\ in the
decomposition tree necessarily leads any attack back to a vertex it
has visited (broken into) already, after such removal the arcs between
components form a directed tree as in Figure~\ref{fig:4ALlevels123}
(a). Each non-root component $\component_i$ (\eg, $C_3$) has exactly
one parent component $\component$ in the cleaned-up tree (\eg,
$C_1$). The respective subnetwork $N \in \component$ (\eg, $N_3$) is a
cut vertex in $LN$. Thus, as claimed above, $N$ is the \emph{only}
vertex, in $LN$, that connects into $\component_i$.

Obviously, all attacks on $\component_i$ must pass through its parent
$N$. Further, the vertices and arcs removed by clean-up cannot be part
of an optimal attack. Thus Level~1 is loss-free. To state this---and
the other properties of 4AL---formally, we need some notations. We
will use $V^*$ to denote the real (optimal) expected value of an
attack, and $V$ to denote the 4AL approximation. The attacked object
is given as the argument. For example, $V^*(LN)$ is the expected value
of attacking $LN$; $V(C,\pivotingrewardfn)$ is the outcome of running
4AL Level~2 on component $\component$ and pivoting reward function
$\pivotingrewardfn$.

\begin{proposition}\label{pro:level1correct} 
Let $LN$ be a logical network. Say that, for all calls to 4AL Level~2
made by 4AL Level~1 when run on $LN$, we have $V(C,\pivotingrewardfn)
= V^*(C,\pivotingrewardfn)$. Then $V(LN) = V^*(LN)$. If
$V(C,\pivotingrewardfn) \leq V^*(C,\pivotingrewardfn)$ for all calls
to 4AL Level~2, then $V(LN) \leq V^*(LN)$.
\end{proposition}

Consider now Algorithm~\ref{alg:level2}. Our previous description was
imprecise in omitting the additional algorithm argument
$\pivotingrewardfn$. This integrates with the algorithm by being
passed on, for every subnetwork on the paths we consider (line 7), to
Algorithm~\ref{alg:level3} which adds it to the reward obtained for
hacking into that subnetwork (Algorithm~\ref{alg:level3} line 4).

$\reward$ aggregates the values (lines 1, 9), over all rewarded
subnetworks $N$. This aggregation is made conservative by removing all
rewards---pivoting rewards as well as the own rewards of the
individual machines involved---that have already been accounted for
(line 10). Regarding the individual machines,
Algorithm~\ref{alg:level2} uses the shorthands (a) $\rewardfn(N) > 0$
(line 2) and (b) $\rewardfn(N) \leftarrow 0$ (line 10); (a) means that
there exists $m \in N$ so that $\rewardfn(m) > 0$; (b) means that
$\rewardfn(m) \leftarrow 0$ for all $m \in N$. Regarding pivoting
rewards, note that line 10 of Algorithm~\ref{alg:level2} modifies the
function $\pivotingrewardfn$ maintained by
Algorithm~\ref{alg:level1}. This does not lead to conflicts because,
at the time when Algorithm~\ref{alg:level1} calls
Algorithm~\ref{alg:level2} on component $C$, all descendants of $C$ in
$LN$ have already been processed, and thus in particular
Algorithm~\ref{alg:level1} will make no further updates to the value
of $\pivotingrewardfn(N)$, for any $N \in C$.

By $\component_\start$ (line 4) we denote the set $\{N \in \component
\mid \exists N' \in LN, N' \not \in \component: (N',N) \in LN\}$ of
subnetworks that serve as an entry into $\component$ (\eg, $N_1$ and
$N_3$ for $C_1$ in Figure~\ref{fig:4ALlevels123} (b)).  Note in line 4
that the path $P$ starts with a firewall $F_0$. To understand this,
consider the situation addressed. The algorithm assumes that the
parent $N$ of $\component$ (\start, for component $C_1$ in
Figure~\ref{fig:4ALlevels123} (b)) is under control. But then, to
break into $\component$, we still need to traverse an arc from $N$
into $\component$. $F_0$ is the firewall on the arc chosen by $P$
($F^*_1$ or $F^*_3$ in Figure~\ref{fig:4ALlevels123} (b)).

The calls to Level~3 (line 7) comprise the network $N_i$ to be hacked
into, the firewall $F_{i-1}$ that must be traversed for doing so, the
pivoting reward of $N_i$, as well as the ongoing path reward
$\reward(P)$ which gets propagated backwards along the path. Clearly,
this is equivalent to the sequence of attacks required to execute $P$,
and harvesting all pivoting rewards associated with such an
attack. Thus, with the conservativeness of the aggregation across the
subnetworks $N$, we get:

\begin{proposition}\label{pro:level2conservative} 
Let $\component$ be a biconnected component, and let
$\pivotingrewardfn$ be a pivoting reward function. Say that, for all
calls to 4AL Level~3 made by 4AL Level~2 when run on
$(\component,\pivotingrewardfn)$, we have
$V(F,N,\pivotingreward,\pathreward) \leq
V^*(F,N,\pivotingreward,\pathreward)$. Then $V(C,\pivotingrewardfn)
\leq V^*(C,\pivotingrewardfn)$.
\end{proposition}

Algorithms~\ref{alg:level3} and~\ref{alg:level4} should be
self-explanatory, given our previous discussion. Just note that the
pivoting reward $\pivotingreward$ is represented by the arc from $m$
to $C_3$ in Figure~\ref{fig:4ALlevels123} (c), which is accounted for
by simply adding it to the value of $m$ (Algorithm~\ref{alg:level3}
line 4). The path reward $\pathreward$ (not illustrated in
Figure~\ref{fig:4ALlevels123} (c)) is also added to the value of $m$
(Algorithm~\ref{alg:level3} line 4). Max'ing over attacks on the
individual machines $m$ is, obviously, a conservative approximation
because attack strategies are free to choose $m$. Thus:

\begin{proposition}\label{pro:level3conservative} 
Let $F$ be a firewall, let $N$ be a subnetwork, let $\pivotingreward$
be a pivoting reward, and let $\pathreward$ be a path reward. Say
that, for all calls to 4AL Level~4 made by 4AL Level~3 when run on
$(F,N,\pivotingreward,\pathreward)$, we have $V(F,m,\reward) \leq
V^*(F,m,\reward)$. Then $V(F,N,\pivotingreward,\pathreward) \leq
V^*(F,N,\pivotingreward,\pathreward)$.
\end{proposition}

\subsection{Policy Construction}

At Level~1, the global policy is constructed from the Level~2 policies
simply by following the tree decomposition: starting at the tree root,
we execute the Level~2 policies for all reached components (in any
order); once a hack into a component succeeds, the respective children
components become reached. At Level~2, i.e., within a bi-connected
component $C$, the policy corresponds to the set of paths $P$
considered by Algorithm~\ref{alg:level2}. Each $P$ is processed in
turn. For each node $N$ in $P$ (until failure to enter that
subnetwork), we call the corresponding Level~3 policy.

At Level~3, i.e., considering a single subnetwork $N$, our policy
simply attacks the machine $m \in N$ that yielded the maximum in
Algorithm~\ref{alg:level3}. The policy first attacks $m$ through the
firewall, using the respective Level~4 policy. In case the attack
succeeds, the policy attacks the remaining machines $m' \in N$ in any
order (i.e., for each $m'$, we perform the associated Level~4 policy
until termination). At Level~4, the policy is the POMDP policy
returned by our POMDP solver.

\ifthenelse{\isundefined{\jfpda}}{
\begin{figure*}[t]
\begin{tabular}{cc}
\includegraphics[width=\columnwidth]{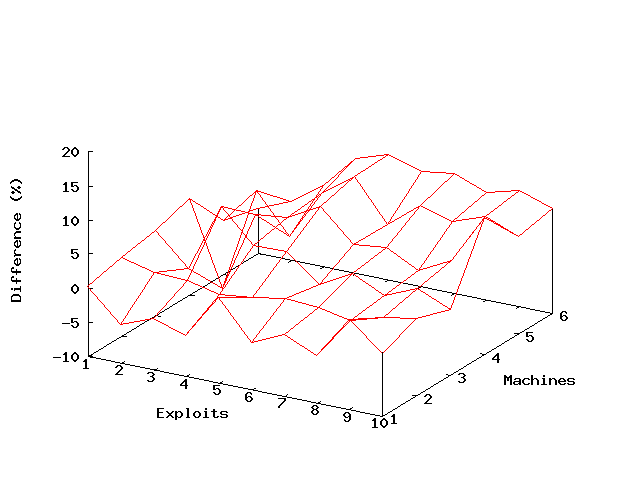} &
\includegraphics[width=\columnwidth]{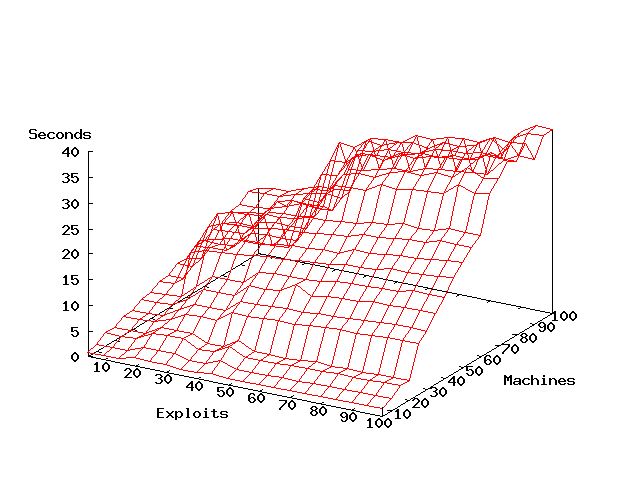}\\[-0.6cm]
(a) Attack quality comparison.  & (b) Runtime of 4AL. 
\end{tabular}
\vspace{-0.2cm}
\caption{Empirical results for 4AL compared to a global POMDP model.}
  \label{fig:experiments}
  \vspace{-0.5cm}
\end{figure*}
}{
  \begin{figure}[t]
    \begin{tabular}{cc}
      \includegraphics[width=0.5\linewidth]{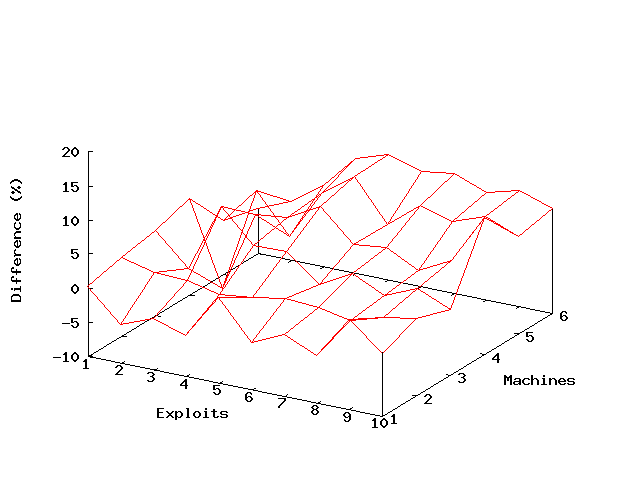} &
      \includegraphics[width=0.5\linewidth]{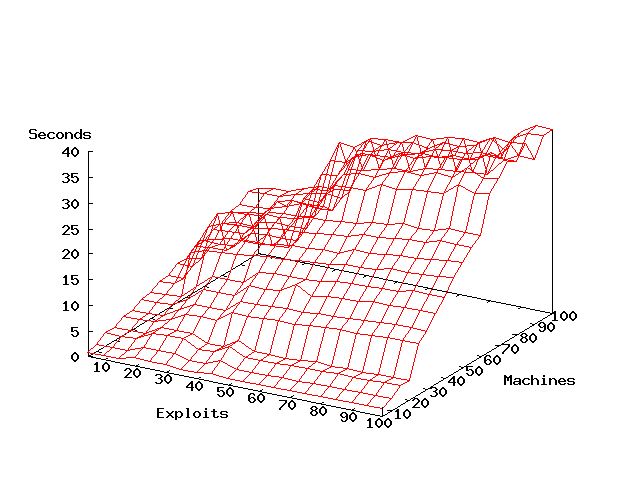}\\[-0.6cm]
      (a) Attack quality comparison.  & (b) Runtime of 4AL. 
    \end{tabular}
    \caption{Empirical results for 4AL compared to a global POMDP model.}
    \label{fig:experiments}
  \end{figure}
}

\section{Experiments}
\label{experiments}

We evaluated 4AL against the ``global'' POMDP model, encoding the
entire attack problem into a single POMDP. The experiments are run on
a machine with an Intel Core2 Duo CPU at 2.2 GHz and 3 GB of RAM.  The
4AL algorithm is implemented in Python.  To solve and evaluate the
POMDPs generated by Level 4, we use the APPL toolkit.\footnote{APPL
  0.93 at http://bigbird.comp.nus.edu.sg/pmwiki/farm/appl/}

\subsection{Test Scenario}
\label{experiments:test-suite}

\begin{figure}[htb]
\vspace{-0.0cm}
\centering
\includegraphics[width=0.95\columnwidth]{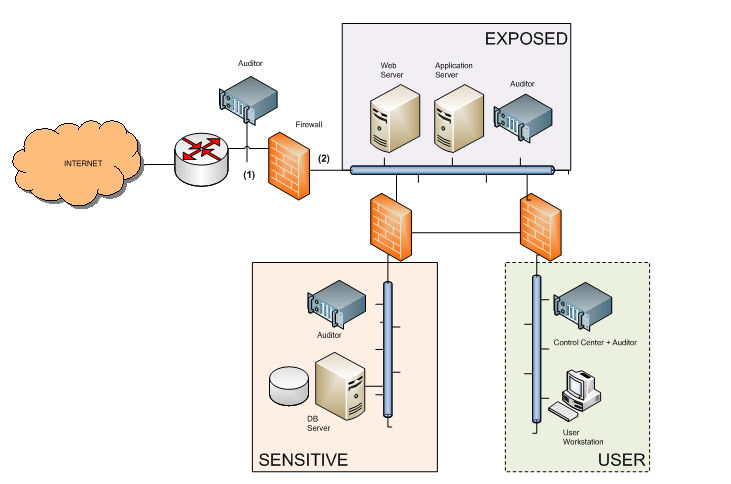}
  \vspace{-0.3cm}
\caption{Network structure in our test suite.}
  \label{fig:scenario}
  \vspace{-0.6cm}
\end{figure}

Our test scenario is based on the network structure shown in
Figure~\ref{fig:scenario}. The attack begins from the Internet
(\start\ is the cloud in the top left corner). The network consists of
three areas---\emph{Exposed, Sensitive, User}---separated by
firewalls. Internally, each of Exposed and Sensitive is fully
connected (i.e., these areas are subnetworks), whereas User consists
of a tree of subnetworks separated by empty firewalls. Only two
machines are rewarded, one in Sensitive (reward 9000) and one in a
leaf subnetwork of User (reward 5000). The cost of port scans and
exploits is 10, the cost of OS detection is 50. We allow to scale the
number of machines $|M|$ by distributing, of every 40 machines, the
first one to Exposed, the second one to Sensitive, and the remaining
38 to User. The exploits are taken from Core Security's database. The
number of exploits $|E|$ is scaled by distributing these over 13
templates, and assigning to each machine $m$ one such template as
$I(m)$ (the known configuration at the time of the last pentest). The
initial belief $b_0(I,T)$, where $T$ is the time elapsed since the
last pentest, is then generated as outlined.

The fixed parameters here (rewards, action costs, distribution of
machines over areas, number of templates) are estimated based on
practical experiences at Core Security. The network structure and
exploits are realistic, and are used for industrial testing in that
company. The main weakness of the scenario lies in the approximation
of software updates underlying $b_0(I,T)$. Altogether, the scenario is
still simplified, but is natural and does approach the complexity of
real-world penetration testing.

For lack of space, in what follows we scale only $|M|$ and $|E|$,
fixing $|T| = 50$. The latter is realistic but challenging: pentesting
is typically performed about once a month; smaller $T$ are easier to
solve as there is less uncertainty.

\subsection{Approximation Loss}
\label{experiments:approximation}

Figure~\ref{fig:experiments} (a) shows the relative loss of quality
when running 4AL instead of a global POMDP solution, for values of
$|E|$ and $|M|$ where the latter is feasible. %
We show
$\mathit{quality}(\mathit{global\mbox{-}POMDP})-\mathit{quality}(\mathit{4AL})$
in percent of $\mathit{quality}(\mathit{global\mbox{-}POMDP})$. Policy
quality here is estimated by running 2000 simulations. That
measurement incurs a variance, which is almost stronger than the very
small quality advantage of the global POMDP solution. The maximal loss
for any combination of $|E|$ and $|M|$ is $14.1\%$ (at $|E|=7$,
$|M|=6$), the average loss over all combinations is $1.96\%$. The
average loss grows monotonically over $|M|$, from $-1.14\%$ for
$|M|=1$ to $4.37\%$ for $|M|=6$. Over $|E|$, the behavior is less
regular; the maximum average loss, $5.4\%$, is obtained when fixing
$|E|=5$.

\subsection{Scaling Up}
\label{experiments:scaling}

Figure~\ref{fig:experiments} (b) shows the runtime of 4AL when scaling
up to much larger values of $|E|$ and $|M|$. The scaling behavior over
$|M|$ clearly reflects the fact that 4AL is polynomial in that
parameter, except for the size of biconnected components (which is $3$
here). Scaling $E$ yields more challenging single-machine POMDPs,
resulting in a sometimes steep growth of runtime. However, even with
$|M|$ and $|E|$ both around $100$, which is a realistic size in
practice, the runtime is always below $37$ seconds.

\section{Conclusion}
\label{conclusion}

We have devised a POMDP model of penetration testing that allows to
naturally represent many of the features of this application, in
particular incomplete knowledge about the network configuration, as
well as dependencies between different attack possibilities, and
firewalls. Unlike any previous methods, the approach is able to
intelligently mix scans with exploits. While this accurate solution
does not scale, large networks can be tackled by a decomposition
algorithm. Our present empirical results suggest that this is
accomplished at a small loss in quality relative to a global POMDP
solution.

An important open question is to what extent our POMDP + decomposition
approach is more cost-effective than the classical planning solution
currently employed by Core Security. Our next step will be to answer
this question experimentally, comparing the attack quality of 4AL
against that of the policy that runs extensive scans and then attaches
FF's plan for the most probable configuration.

4AL is a domain-specific algorithm and, as such, does not contribute
to the solution of POMDPs in general. At a high level of abstraction,
its idea can be understood as imposing a template on the policy
constructed, thus restricting the space of policies explored (and
employing special-purpose algorithms within each part of the
template). In this, the approach is somewhat similar to known POMDP
decomposition approaches (e.g.,
\cite{PinGorThr-uai03,mueller:biundo:ki-11}). It remains to be seen
whether this connection can turn out fruitful for either future work
on attack planning, or POMDP solving more generally.

The main directions for future work are to devise more accurate models
of software updates (hence obtaining more realistic designs of the
initial belief); to tailor POMDP solvers to this particular kind of
problem, which has certain special features, in particular the absence
of non-deterministic actions and that some of the uncertain parts of
the state (e.g.\ the operating systems) are static; and to drive the
industrial application of this technology. We hope that these will
inspire other researchers as well.

\medskip
\noindent
\textbf{Acknowledgments.} Work performed while J\"org Hoffmann was
employed by INRIA, Nancy, France.

\bibliographystyle{aaai}
\bibliography{../../biblio/planning}

\pseudocode{

\newpage

\begin{appendix}
\input{pseudocode}
\end{appendix}
}

\end{document}